\documentclass{article} % For LaTeX2e
\usepackage[final]{colm2026_conference}

\usepackage[utf8]{inputenc} % allow utf-8 input
\usepackage[T1]{fontenc}    % use 8-bit T1 fonts
\usepackage{microtype}      % microtypography
\usepackage{hyperref}       % hyperlinks
\usepackage{url}            % simple URL typesetting
\usepackage{booktabs}       % professional-quality tables
\usepackage{amsfonts}       % blackboard math symbols
\usepackage{nicefrac}       % compact symbols for 1/2, etc.
\usepackage{xcolor}         % colors

\usepackage{amsmath,amsfonts,bm}

\def\eqref#1{equation~\ref{#1}}
\def\1{\bm{1}}

\DeclareMathAlphabet{\mathsfit}{\encodingdefault}{\sfdefault}{m}{sl}
\SetMathAlphabet{\mathsfit}{bold}{\encodingdefault}{\sfdefault}{bx}{n}

\usepackage{amssymb}
\usepackage{algorithm}
\usepackage{algpseudocode}
\usepackage{graphicx}
\usepackage{xspace}
\usepackage{subcaption}
\usepackage{lineno}         % line numbers for submission (required by COLM)

\definecolor{darkblue}{rgb}{0, 0, 0.5}
\hypersetup{colorlinks=true, citecolor=darkblue, linkcolor=darkblue, urlcolor=darkblue}

\newcommand{\solver}{\texttt{dp\_knapsack\_sliding\_hirschberg}\xspace}
\newcommand{\dpfull}{\texttt{dp\_knapsack}\xspace}
\newcommand{\ilp}{\texttt{ilp\_knapsack}\xspace}
\newcommand{\greedy}{\texttt{greedy\_knapsack}\xspace}
\newcommand{\mincut}{\texttt{min\_cut\_re\allowbreak materi\allowbreak alization\_\allowbreak partition}}

\title{Memory-Efficient Activation Checkpointing\\
with Sliding Window and Hirschberg's Algorithm\\for 0/1 Knapsack Solving in PyTorch}

\author{%
  Jędrzej Maczan \\
  Cohere Labs Community \\
  Poland \\
  \texttt{jedrzej@maczan.pl} \\
}

\begin{document}

\ifcolmsubmission
\linenumbers
\fi

\maketitle

\begin{abstract}
Activation checkpointing minimizes the runtime of neural networks under a given memory budget, by selecting which intermediate tensors to store and which to recompute. PyTorch solves this as a 0/1 knapsack problem, where operations from a joint forward-backward computation graph are items with a memory cost (weight) and a runtime saving (value). The default solver, \dpfull, allocates a full dynamic programming (DP) table of shape $(n+1) \times (W+1)$, where $n$ is the number of operations and $W$ is the quantized memory budget. This method is resource-hungry and crashes at $n = 100$ items on a machine with 64\,GB RAM.

In this paper, we introduce \solver, which combines the sliding window trick and Hirschberg's algorithm to reduce peak memory from $O(nW)$ to $O(W)$ while preserving the exact optimal solution. Our experiments show successful knapsack execution at $n = 2{,}000$ (peak 58.4\,GB), where \dpfull fails at $n = 100$, a \textbf{20$\times$ increase in computable problem size}. In addition, our benchmarks show a consistent \textbf{25--28\% runtime speedup} over \dpfull.

The implementation is merged into PyTorch and released in version 2.10.
\end{abstract}

\section{Introduction}

\emph{Activation checkpointing}~\citep{chen2016training} is a method of minimizing program runtime. PyTorch~\citep{paszke2019pytorch} implements this in \texttt{torch.compile}, where it captures the graph~\citep{reed2022torchfx} and uses \mincut to chooses which operations to save or recompute given a budget~\citep{pytorch2024actckpt}. Algorithmically, this is a 0/1 knapsack problem, where every candidate operation is an \emph{item} with a memory cost (weight) and a runtime saving (value), and the goal is to maximize the total savings within a given capacity.

The default solver, \dpfull, uses standard DP with a full 2D table of shape $(n+1)\times(W+1)$, where $n$ is the number of operations and $W$ the quantized memory budget. We empirically show that problems with as few as $n = 100$ items result with out-of-memory (OOM) crashes on 64\,GB hardware. We introduce \solver, which combines two algorithmic improvements over the default solver: (i)~a \textbf{sliding window} that allows to store only two rows of the DP table, thus reduces memory usage from $O(nW)$ to $O(W)$, and (ii)~\textbf{Hirschberg algorithm}, which recovers the optimal item selection from these two rows. Their application to PyTorch was suggested in a TODO comment in the original \dpfull source code. Our contribution is successful combination of sliding window and Hirschberg algorithm as a knapsack solver in PyTorch, implementation shipped in production to the machine learning community in PyTorch 2.10 and an empirical evaluation of the memory and runtime gains over the default solver.

\section{Background}

\paragraph{0/1 knapsack and dynamic programming.}
Given $n$ items with weights $w_i$ and values $v_i$ and a capacity $W$, the we search for such $S\subseteq[n]$ with $\sum_{i\in S} w_i \leq W$ that they maximize $\sum_{i\in S} v_i$. A classical DP solution~\citep{bellman1957dynamic} builds a full table $T[i][c]$ - best value using items $1,\dots,i$ under capacity $c$ - ~\citep{kellerer2004knapsack, martello1990knapsack, pisinger1997minimal} with $T[i][c] = \max(T[i-1][c],\, T[i-1][c-w_i]+v_i)$ if $w_i\le c$. This approach takes $O(nW)$ time and space and recovering the selected items requires backtracking through all the $n$ rows. Checkpointing originates from reverse-mode automatic differentiation~\citep{griewank1992logarithmic, griewank2000revolve} and PyTorch's \mincut automates the save/recompute decision~\citep{pytorch2024actckpt, he2023transcending}. Earlier work used DP to decide which activations to store under a budget~\citep{gruslys2016memory}. Reversible layers avoid storage entirely~\citep{gomez2017reversible, kitaev2020reformer}.

\paragraph{Sliding window and Hirschberg's algorithm.}
Since only the optimal \emph{value} (the last column and row) is needed in practice for memory planning purposes, two rows of the DP table are enough. This change alone reduces memory use to $O(W)$~\citep{cormen2009algorithms}. We could end at this optimization if we didn't need the items \emph{selection}, but in case of memory planning we obviously need it. \citet{hirschberg1975linear} introduced a divide-and-conquer algorithm that achieves linear memory use while recovering the full and exact solution, trading a constant factor of extra time~\citep{llorens2022hirschberg}. This algorithm splits the items into two halves, computes each half's DP profile in $O(W)$ space using the sliding window, combines the profiles (DP rows) to find the optimal split and recurses. This was adapted to knapsack in competitive programming~\citep{codeforces2016hirschberg} and is the basis for our solver.

\section{Method}

\begin{figure}[t]
\centering
\begin{subfigure}{0.6\linewidth}
  \centering
  \includegraphics[width=\linewidth]{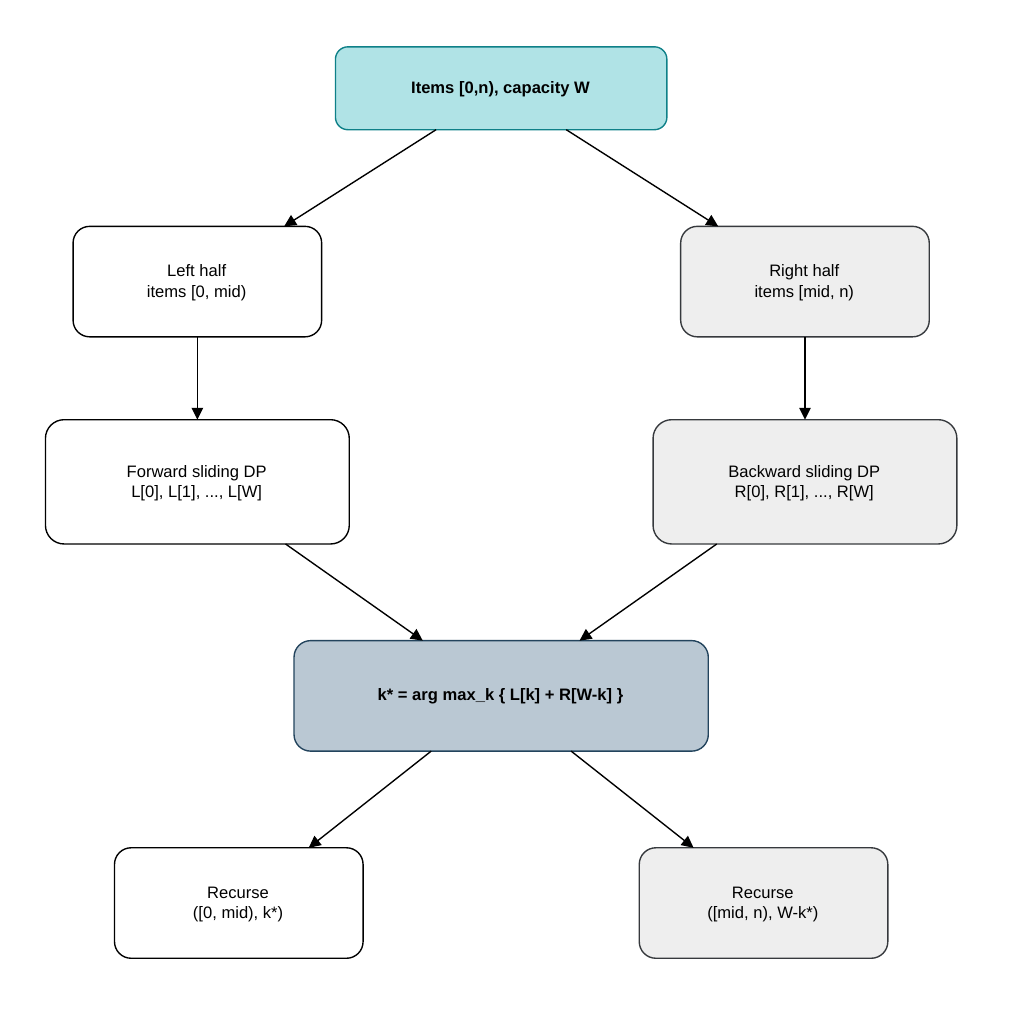}
\end{subfigure}
\caption{DP Hirschberg with sliding-window. The item set is split in the middle. Forward and backward DP passes compute vectors L and R. The optimal split $k\text{*=argmax}_k{L[k]+R[W-k]}$ creates two subproblems.}
\label{fig:algo}
\end{figure}

\solver combines the sliding window with Hirschberg's divide-and-conquer to solve knapsack in $O(W)$ space while recovering the exact selection. To avoid stack overflow for large $n$, it uses an explicit LIFO stack (Algorithm~\ref{alg:main}). Each frame stores an index range $[\ell,r)$ and capacity $c$. A base case ($r-\ell=1$) saves the single item if its weight fits and its value is positive. Zero-weight items are always saved. For all the other cases, the range is split in the middle, a sliding-window DP profile is computed for each half, and the argmax of the combined profiles gives the optimal capacity split $c^*$~(Figure \ref{fig:algo})

\begin{algorithm}[t]
\caption{\solver}
\label{alg:main}
\begin{algorithmic}[1]
\Require Items $(w_i, v_i)_{i=0}^{n-1}$, capacity $W$
\Ensure Lists \textit{saved}, \textit{recomputable}
\State $\textit{saved} \gets []$;\; $\textit{recomputable} \gets []$;\; $\textit{stack} \gets [(0,\; n,\; W)]$
\While{stack is not empty}
    \State $(\ell, r, c) \gets \textit{stack.pop}()$;\quad \textbf{if} $r-\ell = 0$ \textbf{then continue}
    \If{$r - \ell = 1$}
        \State \textbf{if} $w_\ell \leq c$ \textbf{and} $v_\ell > 0$ \textbf{then} $\textit{saved.append}(\ell)$ \textbf{else} $\textit{recomputable.append}(\ell)$
    \Else
        \State $m \gets \ell + \lfloor (r - \ell) / 2 \rfloor$
        \State $P_1 \gets \textsc{SlidingWindowDP}(\ell,m,c)$;\quad $P_2 \gets \textsc{SlidingWindowDP}(m,r,c)$
        \State $c^* \gets \arg\max_{0 \leq k \leq c}\,(P_1[k] + P_2[c - k])$ \Comment{$P_2$ accessed in reverse}
        \State $\textit{stack.push}((m,\; r,\; c - c^*))$;\quad $\textit{stack.push}((\ell,\; m,\; c^*))$
    \EndIf
\EndWhile
\State \Return $\textsc{Sort}(\textit{saved})$,\; $\textsc{Sort}(\textit{recomputable})$
\end{algorithmic}
\end{algorithm}

\textsc{SlidingWindowDP}$(\ell,r,c)$ computes the DP profile for items $\ell,\dots,r-1$ within capacity $c$, using two row buffers of size $c+1$.

\textbf{Space:} four DP buffers of size $W+1$, a stack of $O(\log n)$ frames, and output lists of size $n$, giving $O(W+n)$ peak space versus $O(nW)$ for \dpfull.

\textbf{Time:} each of $O(\log n)$ levels does $O(nW)$ work, so $O(nW\log n)$ versus $O(nW)$ (see Section~\ref{sec:eval}).

\section{Evaluation}
\label{sec:eval}

\begin{figure}[t]
\centering
\begin{subfigure}{0.49\linewidth}
  \centering
  \includegraphics[width=\linewidth]{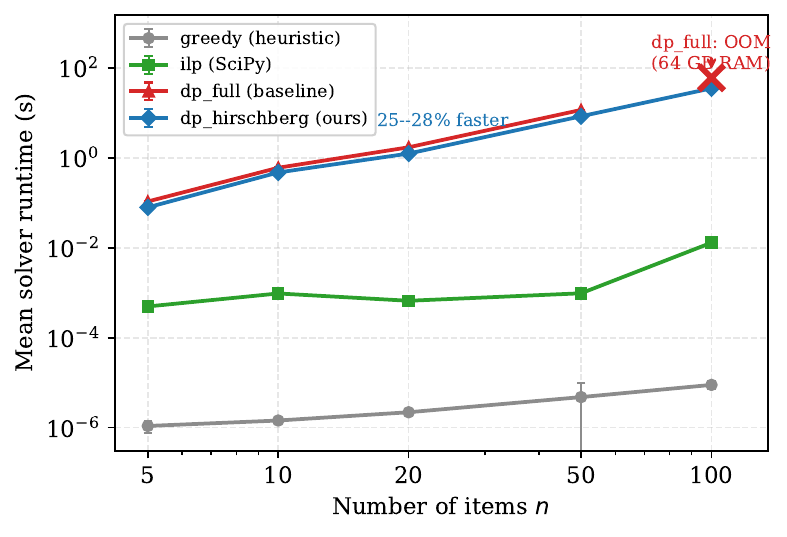}
\end{subfigure}
\hfill
\begin{subfigure}{0.49\linewidth}
  \centering
  \includegraphics[width=\linewidth]{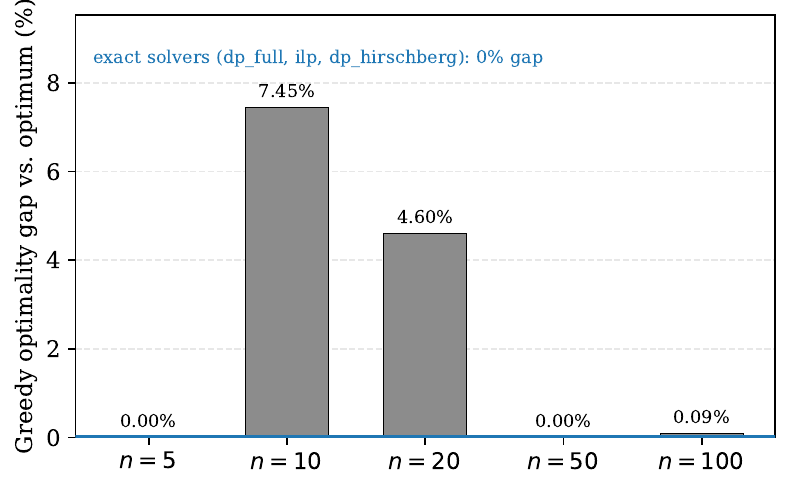}
\end{subfigure}
\caption{\textbf{Left:} mean solver runtime versus $n$. \textbf{Right:} \greedy leaves an instance-dependent optimality gap of up to 7.4\%, whereas the exact solvers (\dpfull, \ilp, \solver) all reach the optimum.}
\label{fig:bench}
\end{figure}

We benchmark all four PyTorch knapsack solvers on synthetic instances mirroring real activation memory planning ($W$ from $1.4\times10^8$ at $n{=}1$ to $3.8\times10^8$ at $n{=}100$), each averaged over 1{,}000 runs on an idle machine (Ubuntu 24.04.2, AMD Ryzen 7 9800X3D, 64\,GB RAM). Table~\ref{tab:runtime} and Figure~\ref{fig:bench} show runtime and exactness. \solver is consistently 25--28\% faster than PyTorch's default \dpfull (Figure~\ref{fig:bench}, left), \ilp is faster still but requires SciPy dependency, and \greedy is fastest but suboptimal.

\dpfull allocates a $(n+1)\times(W+1)$ table: at $n{=}100$, $W\approx3.8\times10^8$ this is $\approx304$\,GB, and it fails with an out-of-memory error on a 64\,GB machine. \solver uses a fixed number of row buffers and runs at $n = 2{,}000$ with a 58.4\,GB peak on the same machine, resulting with a \textbf{20$\times$ increase in computable problem size}, with the DP table shrinking from $\sim$304\,GB to $\sim$6\,GB at $n{=}100$.

The three solvers (\dpfull, \ilp, \solver) produce the exact, optimal solution at every size, while \greedy produces up to 7.4\% below optimal (Figure~\ref{fig:bench}, right). This difference between \greedy and \solver makes the latter a substantial improvement to PyTorch ecosystem - it keeps \dpfull's exactness while removing its memory bottleneck. Thus, as a rule of thumb we recommend \ilp when SciPy is available, \ilp when exact solutions don't matter and \solver otherwise for large or memory-constrained graphs.

\begin{table}[t]
\centering
\caption{Mean solver runtime in seconds over 1,000 runs}
\label{tab:runtime}
\setlength{\tabcolsep}{4pt}
\small
\begin{tabular}{lrrrrr}
\toprule
Solver & $n=5$ & $n=10$ & $n=20$ & $n=50$ & $n=100$ \\
\midrule
\greedy         & $1.1 \times 10^{-6}$ & $1.4 \times 10^{-6}$ & $2.2 \times 10^{-6}$ & $4.8 \times 10^{-6}$ & $8.9 \times 10^{-6}$ \\
\ilp            & $5.0 \times 10^{-4}$ & $9.6 \times 10^{-4}$ & $6.6 \times 10^{-4}$ & $9.7 \times 10^{-4}$ & $1.3 \times 10^{-2}$ \\
\dpfull         & $0.107$ & $0.602$ & $1.716$ & $11.613$ & OOM \\
OURS  & $0.079$ & $0.473$ & $1.244$ & $8.396$ & $35.10$ \\
\bottomrule
\end{tabular}
\end{table}

\section{Discussion and Related Work}

\dpfull fails when $n$ is large - with long sequence lengths, wide graphs, or full model graph captures in \texttt{torch.compile}~\citep{shoeybi2019megatron, narayanan2021efficient}. \solver has worse asymptotic time than \dpfull ($O(nW\log n)$ vs.\ $O(nW)$)~\citep{llorens2022hirschberg}. Yet it comes with runtime speedup, thanks to using fixed buffers, and optimizing memory-access patterns, which are shown to dominate performance~\citep{dao2022flashattention}. The quality, usefulness in the mainstream deep learning framework and correctness of \solver was reviewed and merged into PyTorch and shipped in PyTorch 2.10~\citep{pytorch210}.

Activation checkpointing~\citep{chen2016training} is now considered as a standard technique in machine learning compilers. Subsequent work explores rematerialization strategies~\citep{kirisame2021dynamic, beaumont2021efficient, schuler2022xengine, kumar2019efficient, kusumoto2019graph}, optimal graph partitioning~\citep{herrmann2019optimal}, and automated systems~\citep{jain2020checkmate}. These are complemented by techniques like optimizer-state sharding, offloading~\citep{rajbhandari2020zero, rasley2020deepspeed, ren2021zerooffload} and selective recomputation~\citep{korthikanti2023reducing}, and are catalogued in broader surveys~\citep{tian2025survey}. The sliding-window reduction is a standard textbook technique~\citep{cormen2009algorithms, kellerer2004knapsack}, and Hirschberg's algorithm~\citep{hirschberg1975linear} was originally proposed for sequence alignment.

\section{Conclusion}

We presented \solver, a memory-efficient exact 0/1 knapsack solver for PyTorch's activation memory planning. It combines the sliding window with Hirschberg's divide-and-conquer algorithm and reduces peak solver memory from $O(nW)$ to $O(W)$, while preserving the optimal solution. It empirically shows up to 20$\times$ increase in computable problem size and a 25--28\% runtime speedup compared to PyTorch default knapsack solver. \solver is merged into PyTorch since 2.10.

\bibliography{main}
\bibliographystyle{colm2026_conference}

\end{document}